\documentclass[runningheads]{llncs}
\usepackage[T1]{fontenc}
\usepackage{graphicx}
\usepackage{booktabs}
\usepackage[table]{xcolor}
\usepackage{preamble}
\usepackage{orcidlink}

\usepackage{hyperref}

\begin{document}
\title{On the Robustness of Vision-Language Models in Zero-shot Privacy Classification}
%
\titlerunning{Robustness of VLMs in Zero-shot Privacy Classification}
%

\author{Alina Elena Baia\inst{1}\orcidlink{0000-0001-5553-776X} \and
Alessio Xompero\inst{2}\orcidlink{0000-0002-8227-8529} \and
Andrea Cavallaro\inst{1}\orcidlink{0000-0001-5086-7858}}

\authorrunning{A.E. Baia et al.}

%
\institute{EPFL, Switzerland
\and Independent Researcher
\\
\email{\{alina.baia,andrea.cavallaro\}@epfl.ch, alessio.xompero@gmail.com} 
}
\maketitle              

\begingroup
\renewcommand\thefootnote{}
\footnotetext{\scriptsize A.~E.~Baia and A.~Xompero equally contributed.}
\endgroup

\begin{abstract}
Automatic systems for document understanding require multimodal models that accurately identify sensitive visual content, even in the presence of image degradations. Instruction-following large Vision-Language Models (VLMs) are expected to generalise across domains and tasks without requiring any specific adaptation. In this work, we systematically analyse whether VLMs can be used reliably for image privacy classification in a zero-shot setup. We evaluate and compare the classification performance of three open-source VLMs against purposely built models on two public, standard benchmarks. We assess robustness to image degradations caused by perturbations such as compression, light variations, and random noise, and analyse inference speed and parameter count to deploy VLMs in privacy-aware document processing pipelines. Our results show that large VLMs are robust to input perturbations but are less accurate (and much slower) than smaller privacy models. Scaling alone is not sufficient,  highlighting the advantages of models specifically designed  for privacy classification. 

\end{abstract}
\section{Introduction}
\label{sec:intro}

\setcounter{footnote}{0}

Documents may expose sensitive visual content, such as financial records, and personally identifiable information or identifiable people. Detecting such information is essential for privacy-preserving document redaction and privacy-aware document analysis. 
In this context, identifying whether a document with images contains private information can be framed as an image privacy classification task~\cite{Zerr2012CIKM_PicAlert,Zhao2022ICWSM_PrivacyAlert,Tonge2016AAAI,Tonge2018AAAI,Yang2020PR, Yang2022AAAI,Xompero2025PoPETs}. 

\begin{figure}[t!]
    \centering 
    \scriptsize
    \setlength\tabcolsep{2pt}
    \includegraphics[width=0.75\columnwidth]{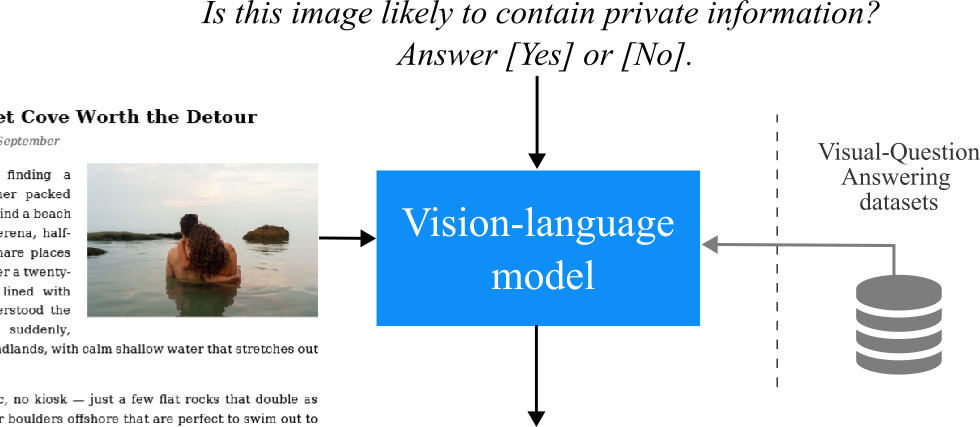}\\
    \begin{tabular}{c}
    \parbox{0.75\columnwidth}{\centering \textbf{LLaVA}~\cite{Liu2023_NeurIPS_LLaVA,Liu2024_CVPR_LLaVa}:``Yes''}\\
      \parbox{0.99\columnwidth}{\centering\textbf{Phi-3-V}~\cite{Abdin2024_ArXiv_Phi3}:``Unable to determine from the image provided''} \\ 
     \parbox{0.80\columnwidth}{\centering\textbf{MiniCPM}~\cite{Yao2024_ArXiv_MiniCPM_V}:``No, the image is not likely to contain private information. It depicts a couple in a natural setting, engaged in an activity that is typically associated with relaxation and intimacy, \ldots ''} \\ 
    \end{tabular}    
    \caption{Zero-shot image privacy classification with a general-purpose pre-trained VLM. Answers from selected open-source VLMs for an image from a public dataset~\cite{Zhao2022ICWSM_PrivacyAlert}.}
    \label{fig:problem}
\end{figure}

Classifying an image as private is challenging due to ambiguous content and subjective preferences~\cite{Orekondy2017ICCV,Baia2024_ECCV_eXCV_Workshop,Xompero2025PoPETs,AriasCabarcos2023PoPETs}. Vision-language models (VLMs)~\cite{Liu2024_CVPR_LLaVa, Abdin2024_ArXiv_Phi3} combine natural-language instructions and image information for classification. In zero-shot image privacy classification (Figure~\ref{fig:problem}), the model predicts image privacy without any fine-tuning for the task. We are interested in quantifying the robustness of VLMs for privacy classification and whether VLMs can outperform specialized vision-only or other multi-modal models~\cite{Orekondy2017ICCV,Baia2024_ECCV_eXCV_Workshop,Xompero2025PoPETs}.

In this paper, we assess the robustness of three instruction-following VLMs for zero-shot image privacy classification across two public datasets, namely PrivacyAlert~\cite{Zhao2022ICWSM_PrivacyAlert} and IPD~\cite{Yang2020PR}. 
As performance criteria, in addition to robustness, we consider accuracy and efficiency. These criteria capture whether a model is robust under common image perturbations, can detect private content and operate under latency and cost constraints. We  analyse robustness to image perturbations and computational trade-off to assess the practicality of general-purpose VLMs compared to methods designed or fine-tuned for image privacy classification. The main findings of this paper for the three open-source instruction-following VLMs in privacy classification are:
\begin{itemize}
    \item VLMs are robust to image distortions,
    \item augmenting the VLM instructions with contextual information and a task-aligned role does not yield accuracy gains, and
    \item a specialised vision-only model with fewer than 50M parameters outperforms the VLMs while offering a hundredfold faster inference speed.
\end{itemize}

\section{Previous Work}

Previous studies evaluated VLMs in recognizing sensitive inputs under a visual question answering paradigm~\cite{Zhang2024_ArXiv_MP2A,zhang2024multitrust,zhang2025reval,samson2024little, Baia2024_ECCV_eXCV_Workshop}. For example, Multi-P2A~\cite{Zhang2024_ArXiv_MP2A}, MultiTrust~\cite{zhang2024multitrust}, and REVAL~\cite{zhang2025reval} categorised privacy tasks in recognizing sensitive inputs, including privacy image recognition and privacy question detection ({\em privacy awareness}), and disclosing sensitive information ({\em privacy violation}). 
These works~\cite{Zhang2024_ArXiv_MP2A, zhang2024multitrust, zhang2025reval} evaluated VLMs either focusing on a subset of images from standard image privacy datasets or using custom datasets that cover limited private aspects such as 
credit card number, home address, email address, phone numbers, and license plates.
Priv-IQ~\cite{priv-iq} evaluated multimodal models across different privacy tasks, including their ability to identify privacy risk from images, and aggregated subsets of data from multiple sources. 
None of the prior works performed evaluations using (whole) standard datasets.

PRIVBENCH introduced a GDPR-aligned benchmark with explicit privacy categories~\cite{samson2024little}, and showed that fine-tuning VLMs on small, high-quality instruction-tuning data (i.e. PRIVTUNE) improves the model's privacy awareness. Similarly, Tsaprazlis et al.~\cite{tsaprazlis2025assessing} report improvements from supervised fine-tuning for privacy-risk detection in images. Despite providing results on standard image privacy datasets, such as PrivacyAlert~\cite{Zhao2022ICWSM_PrivacyAlert} and VISPR~\cite{Orekondy2017ICCV}, comparisons are limited to their own fine-tuned model and omit broader comparisons with prior purpose-built privacy classifiers, resulting in poorly contextualized performance gains relative to established baselines.

Zero-shot image privacy classification was evaluated on PrivacyAlert~\cite{Baia2024_ECCV_eXCV_Workshop}. However, VLMs often generate complex, multi-sentence responses to a privacy query, instead of yes/no answers. The authors therefore discarded such responses in their evaluation process~\cite{Baia2024_ECCV_eXCV_Workshop}, making their work hard to compare fairly with other privacy classification approaches.

Finally, none of these related works~\cite{zhang2025reval, zhang2024multitrust, Zhang2024_ArXiv_MP2A, samson2024little, Baia2024_ECCV_eXCV_Workshop} assessed the robustness of VLMs to image perturbations.
This is important for document image understanding, where documents may undergo degradations due to compression and varying capturing conditions. For example, images may be \textit{compressed} to reduce storage size or satisfy uploading requirements, captured under different \textit{illumination} conditions, or affected by scan noise and other visual artifacts, such as blurring and salt-and-pepper noise. Models intended for privacy-oriented document analysis should be robust under such conditions.

\section{Evaluation Protocol}

We identify VLMs, datasets, and performance measures to use for a systematic evaluation of zero-shot privacy classifiers. We discuss and assess the prompt to use across models and images, and the answer-to-decisions (binary) conversion.

\subsection{VLMs, Standard Datasets, Performance measures}

We consider instruction-based VLMs and assess their decision for image privacy classification assuming that only their free-text generated answers are accessible (black-box setting), which represents the more realistic scenario.
We select Phi-3.5-Vision (Phi-3-V)~\cite{Abdin2024_ArXiv_Phi3}, Large Language and Vision Assistant (LLaVA)~\cite{Liu2023_NeurIPS_LLaVA,Liu2024_CVPR_LLaVa}, and MiniCPM-Llama~\cite{Yao2024_ArXiv_MiniCPM_V}, the top-3 performing models\footnote{We use open-source models (default parameter values) that can run locally to preserve the privacy of the images, enable the assessment of inference speed and hardware resources, and favour reproducibility.} on a recent benchmark for evaluating privacy risks of large VLMs~\cite{Zhang2024_ArXiv_MP2A}. 

To favour continuity with prior works on image privacy classification~\cite{Xompero2025PoPETs,Yang2020PR,Zhao2022ICWSM_PrivacyAlert,Baia2024_ECCV_eXCV_Workshop}, we evaluate the models on the testing sets of standard datasets, such as PrivacyAlert~\cite{Zhao2022ICWSM_PrivacyAlert} (1,796 images) and IPD~\cite{Yang2020PR} (6,912 images). Both datasets have images annotated with public and private labels, and a class imbalance towards the public class~\cite{Xompero2025PoPETs,Zhao2022ICWSM_PrivacyAlert,Yang2020PR}: 25\% and 33\% of the images are labelled as private for PrivacyAlert and IPD, respectively. Note that IPD combines PicAlert~\cite{Zerr2012CIKM_PicAlert}, the first image privacy dataset, with a subset of images from VISPR~\cite{Orekondy2017ICCV} to better balance public and private images, and reduces the bias towards images with people~\cite{Yang2020PR,Xompero2025PoPETs}.

We compute standard classification metrics (reported as percentages): per-class precision, recall, and F1-score, and overall precision, balanced accuracy (average between the recall of the two classes), and accuracy. Because of the class imbalance, we focus on recall of the private class and balanced accuracy for the discussion~\cite{Xompero2025PoPETs}. Moreover, we compute model size (number of trainable parameters) and speed (s/image) to quantify the computational costs of the models and their trade-off with classification performance, and robustness of the models to image perturbations (next sections).

\subsection{Choice of Prompt for Zero-shot Classification}
\label{subsec:taskalignprompt}

We define a first prompt whose question is consistent with the problem formulation~\cite{Xompero2025PoPETs,Zerr2012CIKM_PicAlert,Zhao2022ICWSM_PrivacyAlert} and related benchmarks~\cite{zhang2025reval, zhang2024multitrust}. VLMs are prompt-sensitive~\cite{jin-etal-2022-good, zhang2025reval} and might benefit from more detailed prompts~\cite{zhang2024multitrust, pixels_vs_privacy}. We therefore assess the influence of a second alternative prompt based on annotation instructions.
However, exhaustive prompt analysis is beyond the scope of this work. 

\begin{figure*}[t!]
    \centering 
    \scriptsize
    \setlength\tabcolsep{2.5pt}
    \begin{tabular}{l@{\hspace{-4pt}}ccccc}
    &
    \includegraphics[height=0.11\linewidth]{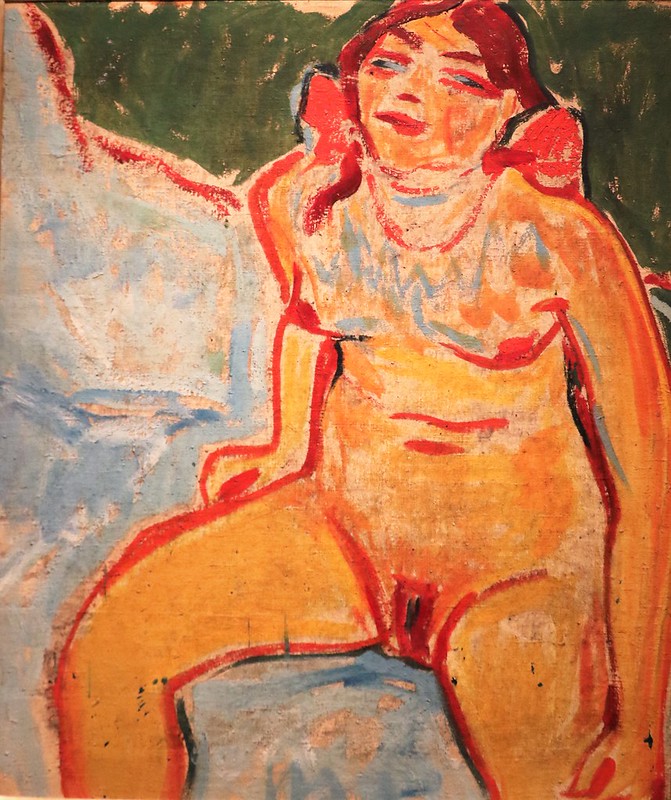} &
    \includegraphics[height=0.11\linewidth]{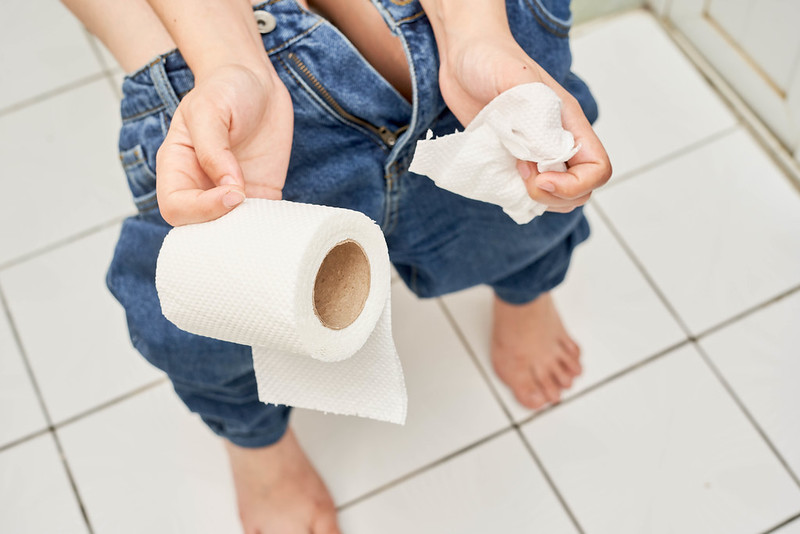} &
    \includegraphics[height=0.11\linewidth]{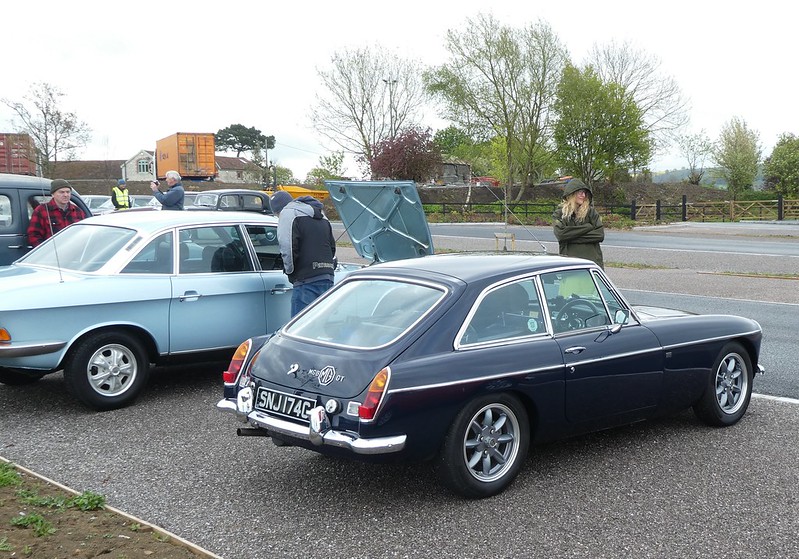} &
    \includegraphics[height=0.11\linewidth]{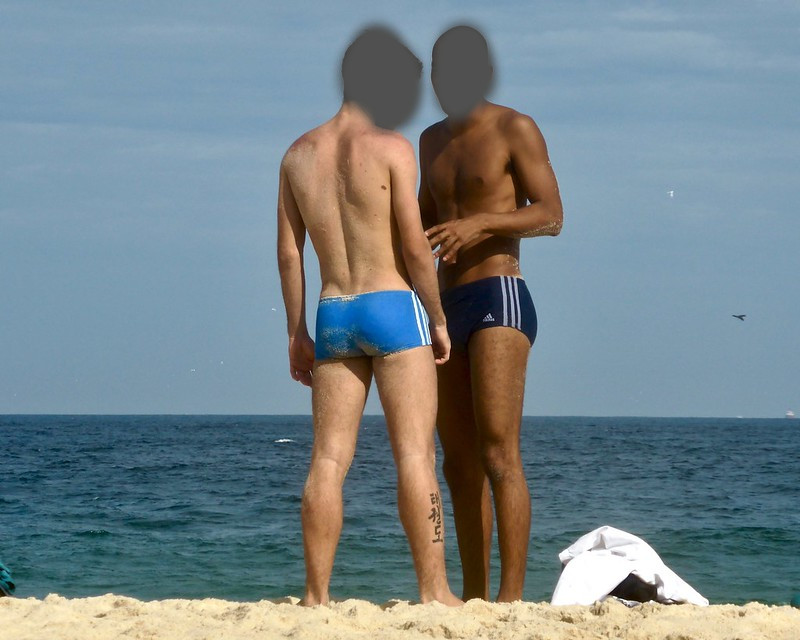} &
    \includegraphics[height=0.11\linewidth]{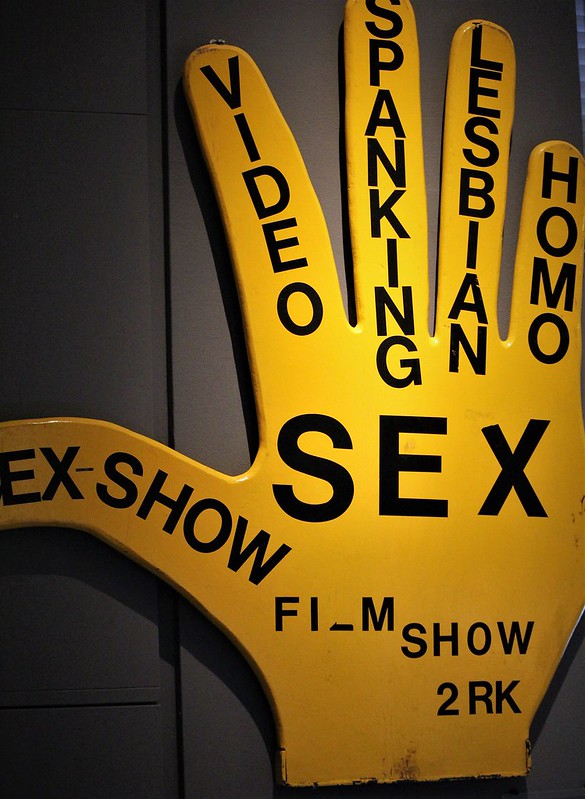} 
    \\
    \toprule
    \textbf{GT:}
        & \textit{Yes} & \textit{Yes} & \textit{No} & \textit{Yes} & \textit{Yes} \\
        \midrule
    \textbf{LLaVA}: &
    ``Yes'' & ``Yes'' & ``Yes'' & ``Yes'' & ``Yes'' \\ 
    \midrule
    \textbf{Phi-3-V}: & 
    \parbox[t]{0.16\linewidth}{``I’m sorry, but I cannot provide details or speculate about the content of the image''} & 
    \parbox[t]{0.16\linewidth}{``No, the image does not contain private information. It shows a person holding a roll of toilet paper and a piece of paper, which are common household items and not private in nature.''} & 
    \parbox[t]{0.16\linewidth}{``No, the image does not contain any visible private information''} & 
    \parbox[t]{0.16\linewidth}{``Unable to determine from the image provided.''} & 
    \parbox[t]{0.16\linewidth}{``No, the image does not contain private information. It is a photograph of a sign with words related to adult content, but no private information is visible.''} \\
    \midrule
    \textbf{MiniCPM}: & ``No'' & ``No''  & \parbox[t]{0.16\linewidth}{``No, the image is not likely to contain private information. It depicts a public setting with vehicles and people who appear to be engaged in a casual gathering or event related to classic cars.\ldots''} & 
    \parbox[t]{0.16\linewidth}{``No, the image is not likely to contain private information. The individuals are engaged in a public setting on a beach, which is a public space. \ldots ''} & ``No'' \\
    \bottomrule

    \end{tabular}
    \caption{VLMs answers to the  prompt  ``\textit{Is this image likely to contain private information? Answer [Yes] or [No].}'' on sample images~\cite{Zhao2022ICWSM_PrivacyAlert}.     KEY -- GT:~ground-truth.}
    \label{fig:vizres}
\end{figure*}

For the \textit{first prompt} (P1), we refer to the Multi-P2A benchmark~\cite{Zhang2024_ArXiv_MP2A}, which pairs each image with a randomly selected input prompt from a predefined set of binary questions. Examples of these prompts are: ``\textit{In your assessment, does this image seem to include private information?  [Yes] or [No].}''; ``\textit{Examine the image for potential privacy concerns and answer [Yes] if it does, [No] if it doesn't.}''; ``\textit{In your assessment, does this image NOT seem to include private information? [Yes] or [No].}''.
These questions are not necessarily framed consistently (yes does not always mean private) or contain wording that can be subjective or opinionated, such as ``in your assessment'' or ``would you consider''. 
To avoid wording such as ``your assessment'', ``examine'', ``potential privacy concerns'', ``risk'', we selected the following prompt for all images and VLMs: 
``\textit{Is this image likely to contain private information? Answer [Yes] or [No].}''. 
Figure~\ref{fig:vizres} shows examples of VLMs answers to this prompt.

For the \textit{second prompt} (P2), we leverage the annotation instructions of PrivacyAlert~\cite{Zhao2022ICWSM_PrivacyAlert}, which include context and a role, thus expected to better guide the model. We derive the following prompt:
``\textit{Assume you have taken these photos, and you are about to upload them on your favourite social network or content sharing site (e.g., Flickr, Facebook, Google+, Instagram). Please tell us whether these images are either private or public in nature. Assume that the people in the photos are those that you know. Private images are images that should be kept confidential for me and selected trusted people only. Public images are ones that anyone in my social network would be OK to see. Answer [Private] or [Public].}'' For this prompt, VLMs are treated as a replacement for the human~\cite{lu2024visionlanguagemodelsreplacehuman, chiang-lee-2023-llms_annotators}.

\subsection{From Model Answers to Privacy Decision}

Despite providing a prompt instructing to answer with a binary choice, VLMs output different types of answers. LLaVA-1.5 outputs answers that are structured according to the instruction, thus allowing a direct automatic conversion into binary values: ``Yes'' or ``Private'' is mapped to Private (1) and ``No'' or ``Public'' is mapped to Public (0).
The models Phi-3-V and MiniCPM do not always provide a simple answer that includes the requested decision. Depending on the image, the answer can be lengthy, complex, and multi-sentence. Examples of these answers include: ``\textit{The image doesn't contain any visible private information.}''; ``\textit{Unable to determine from the image provided.}''; ``\textit{The image appears \ldots not contain any private information \ldots not possible to determine \ldots without additional details about the source or the intended use of the image.}''.

Because of the different formatting, we define the rule-based mapping according to the prompt used and the model's choice of answer. For example, for Phi-3-V, answers can be either ``[Yes]'' or ``Yes'' when using prompt 1, whereas we consider only ``[Yes]'' for MiniCPM. 
If both instructions are present (i.e. ``[Yes]'' and ``[No]'') in the answer, or none, then the rule-based method maps the answer to a lack of decision (ambiguous answer). 

To overcome the lack of decision by the model and still enable a fair comparative evaluation, we provide two extreme approaches as references. One is a \textit{fully permissive} approach that converts the answer into the public choice. The other is a more \textit{fully conservative} or restrictive approach that converts the answer into the private choice.

\subsection{Influence of Prompts and Answer-to-decision Mapping}

Table~\ref{tab:promptscomparison} compares the classification performance of the models when using either of the prompts and either of the reference approaches for Phi-3-V and MiniCPM. LLaVA with P1 achieves the highest balanced accuracy (73.38\% on PrivacyAlert and 71.28\% on IPD) while directly and automatically mapping all answers to the privacy decisions. In contrast, the model with P2 achieves higher accuracy but at the cost of lower balanced accuracy and recall on the private class, almost degenerating into predicting all images as public on IPD.
Similar to LLaVA, Phi-3-V is also sensitive to P2, thus almost degenerating to predict all images as public on IPD. Using P1 together with the mapping to the private class for answers with a clear decision makes Phi-3-V achieve a balanced accuracy of 62.57\% on IPD and 66.80\% on PrivacyAlert. However, the recall lower than 50\% in the private class shows how difficult it is for the model to recognize private images in a zero-shot setting. MiniCPM is highly sensitive to P1 and almost degenerates to predict all images as either public or private, depending on the mapping approach. This shows that the model cannot provide a clear answer for most of the images. However, MiniCPM can provide better decisions when using P2 and the more conservative approach, with a higher balanced accuracy (57.87\% and 67.69\% on IPD and PrivacyAlert, respectively) and recall on the private class (56.90\% and 67.33\% on IPD and PrivacyAlert, respectively). 

\begin{table}[t!]
    \centering
    \scriptsize
    \setlength\tabcolsep{7pt}
    \caption{
    Comparison of VLMs classification results with two text prompts and two reference approaches for resolving answers. KEY --
    AMB: ambiguity resolution, R: recall (private), A: accuracy, BA: Balanced accuracy, PRI:~automatically set to private, PUB:~automatically set to public.
    }
    \begin{tabular}{lcc ccc @{\hskip 12pt} ccc}
    \toprule
     \textbf{Model} & \textbf{Prompt} & \textbf{AMB} & \multicolumn{3}{c}{\textbf{IPD}~\cite{Yang2020PR}} & \multicolumn{3}{c}{\textbf{PrivacyAlert}~\cite{Zhao2022ICWSM_PrivacyAlert}}  \\
    \cmidrule(lr){4-6}\cmidrule(lr){7-9}
    & & & \cellcolor{white!50!}R & \cellcolor{white!50!}BA & A & \cellcolor{white!50!}R & \cellcolor{white!50!}BA & A \\
    \midrule
    \multirow{2}{*}{LLaVA~\cite{Liu2023_NeurIPS_LLaVA,Liu2024_CVPR_LLaVa}}
    & P1 &  -- & 70.49 & 71.28 & 71.54 & 89.33 & 73.38  & 65.42 \\
    & P2 &  -- & 10.16 & 51.99 & 65.93 & 51.56 & 71.54 & 81.51 \\
    \midrule
    \multirow{4}{*}{Phi-3-V~\cite{Abdin2024_ArXiv_Phi3}}
    & \multirow{2}{*}{P1} & PRI  & 31.68 & 62.57 & 72.86  & 42.89 & 66.80 & 78.73 \\
    &  & PUB  & 23.65 & 59.90 & 71.98 & 25.78 & 60.10 & 77.23 \\
    \cmidrule(lr){2-9} 
    & \multirow{2}{*}{P2} & PRI  & 6.42 & 50.64 & 65.38 & 21.78 & 58.66 & 77.06 \\
    &  & PUB  & 6.21 & 50.55 & 65.34 & 16.00 & 55.85 & 75.72 \\
    \midrule
    \multirow{4}{*}{MiniCPM~\cite{Yao2024_ArXiv_MiniCPM_V}}
    & \multirow{2}{*}{P1} & PRI  & 91.19 & 48.49 & 34.26 & 82.44 & 45.12 & 26.50 \\
    &  & PUB   & 1.56 & 50.56 & 66.90 & 5.11 & 52.22 & 75.72\\
    \cmidrule(lr){2-9}
    & \multirow{2}{*}{P2} & PRI  & 56.90 & 57.87 & 58.19 & 67.33 & 67.69 & 67.87 \\
    &  & PUB  & 37.07 & 58.15 & 65.18 & 34.00 & 59.72 & 72.55 \\
    \bottomrule 
    \end{tabular}
    \label{tab:promptscomparison}
\end{table}

We also quantify the percentage of images that the models cannot take a decision and therefore the answers cannot be directly converted into the privacy decision. Phi-3-V cannot provide a decision for 4.47\% and 0.10\% of the images in the testing set of IPD and for 7.07\% and 1.56\% of the images in the testing set of PrivacyAlert, for P1 and P2, respectively. On the contrary, MiniCPM cannot decide for a larger number of images, and especially in the case of P1. MiniCPM cannot provide a clear decision for 92.39\% and 87.97\% of the images in the testing sets of IPD and PrivacyAlert, respectively, when using P1, and 20.21\% and 21.38\% when using P2.

\section{Comparison}

Most of the related methods use images as the only input to learning-based models~\cite{Zerr2012CIKM_PicAlert,Zhao2022ICWSM_PrivacyAlert,Orekondy2017ICCV,Tonge2016AAAI,Tonge2018AAAI,Tonge2020TWEB,Tran2016AAAI_PCNH,Han2022MTA_PrivacyMLMS,Xompero2025PoPETs} or complement and fuse the visual input with other information~\cite{Tonge2018AAAI,Tonge2019WWW}, such as user tags and metadata (e.g. geolocation). 
We compare the 3 VLMs with task-specific uni- and multi-modal models to quantify whether VLMs outperform previous works without any domain adaptation.

\subsection{Privacy Classifiers}

\textit{S2P}~\cite{Xompero2025PoPETs} is a uni-modal model that uses a single image as input and combines transfer learning with a pre-trained Convolutional Neural Network (CNN) to relate privacy with scene types~\cite{Zhou2018TPAMI_Places365}. PCNH~\cite{Tran2016AAAI_PCNH} is a two-branch network that uses one branch (AlexNet pre-trained on ImageNet~\cite{Deng2009CVPR_ImageNet}) for object prediction and another branch for privacy-specific features, and combines their outputs via a late fusion mechanism. Concat~\cite{Tonge2018AAAI} concatenates features from object recognition (CNN pre-trained on ImageNet~\cite{Deng2009CVPR_ImageNet}), scene recognition (CNN pre-trained on Places365~\cite{Zhou2018TPAMI_Places365}), and user tags, followed by a SVM as a privacy classifier. DMFP~\cite{Tonge2019WWW} fuses predictions from different specifically trained classifiers (objects, scenes, user tags) using a weighted majority voting strategy.
P-VilBERT~\cite{Zhao2022ICWSM_PrivacyAlert} fine-tuned a VLM for image privacy using images and tags\footnote{Unlike Zhao et al.~\cite{Zhao2022ICWSM_PrivacyAlert}, we name this model as P-VilBERT to differentiate it from the generic VilBERT and emphasise the fine-tuning for image privacy.
}. GMMF~\cite{Zhao2022ICWSM_PrivacyAlert,Zhao2023TWEB} employs a learnable gating network to dynamically weigh and fuse predictions from uni-modal classifiers (object, scene, user tags).
Privacy VLM~\cite{samson2024little} fine-tuned TinyLLaVA on PRIVTUNE~\cite{samson2024little} to enhance the privacy awareness. 

These methods require task-specific training and rely on specifically designed pipelines~\cite{Tonge2018AAAI,Tonge2020TWEB,Yang2020PR}. Training or fine-tuning these models is difficult because public datasets are limited, severely class-imbalanced towards `public' images, and contain inconsistent or erroneous annotations~\cite{Xompero2025PoPETs,samson2024little}. A summary of the input modalities and training strategies of privacy classifiers is provided in Table~\ref{tab:multimodalcomparison_metadata}.

\subsection{Accuracy and Efficiency}

We compare the classification performance of VLMs against the privacy classifiers (see Table~\ref{tab:multimodalcomparison}) and discuss their accuracy-efficiency relationship on both IPD and PrivacyAlert (see Figure~\ref{fig:sizevsacc}). 
We use the number of parameters for model size and the average speed across 100 images (s/image) for computational cost. Since the models differ substantially in size and hardware requirements, we report inference speed on the available GPUs on which each model could be run to provide a practical deployment-oriented comparison rather than a strictly hardware-controlled benchmark. 
We also consider the best-performing option for Phi-3-V (P1 with conservative approach) and MiniCPM (P2 with conservative approach).

\begin{table*}[t!]
    \centering
    \scriptsize
    \setlength\tabcolsep{7pt}
    \caption{Overview of privacy classifiers in terms of  modalities and training strategy. 
    KEY -- TL:~transfer learning, FT:~fine-tuning, ZS:~zero-shot.}
    \begin{tabular}{l ccccc ccc}
        \toprule
        \multicolumn{1}{l}{\textbf{Method}} & 
        \multicolumn{5}{c}{\textbf{Modalities}} & 
        \multicolumn{3}{c}{\textbf{Training}} \\
        \cmidrule(lr){2-6}\cmidrule(lr){7-9}
        & Objects & Scenes & Tags & Vision & Language & TL & FT & ZS \\
        \midrule

        PCNH~\cite{Tran2016AAAI_PCNH} 
        & \bbox & \wbox & \wbox & \bbox & \wbox & \bbox & \wbox & \wbox \\
        Concat~\cite{Tonge2018AAAI} 
        & \bbox & \bbox & \bbox & \bbox & \wbox & \wbox & \wbox & \wbox \\
        DMFP~\cite{Tonge2019WWW} 
        & \bbox & \bbox & \bbox & \bbox & \wbox & \bbox & \wbox & \wbox \\
        P-VilBERT~\cite{Zhao2022ICWSM_PrivacyAlert,Lu2018NeurIPS_VilBERT} 
        & \wbox & \wbox & \bbox & \bbox & \bbox & \bbox & \bbox & \wbox \\
        GMMF~\cite{Zhao2022ICWSM_PrivacyAlert,Zhao2023TWEB} 
        & \bbox & \bbox & \bbox & \bbox & \wbox & \bbox & \bbox & \wbox \\
        Privacy VLM~\cite{samson2024little} 
        & \wbox & \wbox & \wbox & \bbox & \bbox & \bbox & \bbox & \wbox \\
        S2P~\cite{Xompero2025PoPETs} 
        & \wbox & \bbox & \wbox & \bbox & \wbox & \bbox & \wbox & \wbox \\
        \midrule
        MiniCPM~\cite{Yao2024_ArXiv_MiniCPM_V} 
        & \wbox & \wbox & \wbox & \bbox & \bbox & \wbox & \wbox & \bbox \\
        Phi-3-V~\cite{Abdin2024_ArXiv_Phi3} 
        & \wbox & \wbox & \wbox & \bbox & \bbox & \wbox & \wbox & \bbox \\
        LLaVA~\cite{Liu2023_NeurIPS_LLaVA,Liu2024_CVPR_LLaVa} 
        & \wbox & \wbox & \wbox & \bbox & \bbox & \wbox & \wbox & \bbox \\
         \bottomrule 
    \end{tabular}
    \label{tab:multimodalcomparison_metadata}
\end{table*}
\begin{table*}[t!]
    \centering
    \scriptsize
    \setlength\tabcolsep{3pt}
    \caption{Comparison of image privacy classification results on the testing sets of IPD~\cite{Yang2020PR} and PrivacyAlert (PA)~\cite{Zhao2022ICWSM_PrivacyAlert}. KEY -- P:~precision, R:~recall, F1:~F1-score, A:~accuracy, BA:~balanced accuracy (overall recall); N/A:~not available.}
    \begin{tabular}{ll ccc ccc ccc}
        \toprule
        \multicolumn{1}{l}{\textbf{Dataset}} &  
        \multicolumn{1}{l}{\textbf{Method}} & 
        \multicolumn{3}{c}{\textbf{Private}} & 
        \multicolumn{3}{c}{\textbf{Public}} & 
        \multicolumn{3}{c}{\textbf{Overall}} \\
        \cmidrule(lr){3-5}\cmidrule(lr){6-8}\cmidrule(lr){9-11}
        & & P & \cellcolor{white!50!}R & F1 & P & R & F1 & P & A & \cellcolor{white!50!}BA \\
        \midrule

        \multirow{8}{*}{IPD}
        & All private & 33.33  & 100.00 & 50.00  & 0.00   & 0.00   & 0.00   & 16.67  & 33.33 & 50.00 \\
        & All public  & 0.00   & 0.00   & 0.00   & 66.67  & 100.00 & 80.00  & 33.33  & 66.67 & 50.00 \\ 
        & Random      & 33.68  & 50.61  & 40.44  & 67.01  & 50.17  & 57.38  & 50.35  & 50.32 & 50.39 \\ 
        \cmidrule(lr){2-11}
        & S2P~\cite{Xompero2025PoPETs} 
        & 75.83 & 72.44 & 74.10 & 86.52 & 88.45 & 87.48 & 81.18 & 83.12 & 80.45 \\
        & MiniCPM~\cite{Yao2024_ArXiv_MiniCPM_V} 
        & 40.87 & 56.90 & 47.57 & 73.19 & 58.83 & 65.23 & 57.03 & 58.19 & 57.87 \\ 
        & Phi-3-V~\cite{Abdin2024_ArXiv_Phi3} 
        & 70.74 & 31.68 & 43.76 & 73.23 & 93.45 & 82.11 & 71.98 & 72.86 & 62.57 \\
        & LLaVA~\cite{Liu2023_NeurIPS_LLaVA,Liu2024_CVPR_LLaVa} 
        & 55.79 & 70.49 & 62.28 & 83.00 & 72.07 & 77.15 & 69.40 & 71.54 & 71.28 \\

        \toprule
        \multirow{13}{*}{PA}
        & All private & 25.00 & 100.00 & 40.00 & 0.00  & 0.00   & 0.00  & 12.50 & 25.00 & 50.00 \\
        & All public  & 0.00  & 0.00   & 0.00  & 75.00 & 100.00 & 85.71 & 37.50 & 75.00 & 50.00 \\
        & Random      & 74.27 & 50.67  & 60.24 & 24.23 & 47.33  & 32.05 & 49.25 & 49.83 & 49.00 \\
        \cmidrule(lr){2-11}
        & *PCNH~\cite{Tran2016AAAI_PCNH} 
        & 70.60 & 51.10 & 59.30 & 85.10 & 92.90 & 88.80 & 77.85 & 83.17 & 72.00 \\
        & *Concat~\cite{Tonge2018AAAI} 
        & 62.60 & 71.60 & 66.80 & 90.00 & 85.80 & 87.90 & 76.30 & 82.22 & 78.70 \\
        & *DMFP~\cite{Tonge2019WWW} 
        & 66.60 & 65.60 & 66.10 & 88.60 & 89.00 & 88.80 & 77.60 & 83.17 & 77.30 \\
        & *P-VilBERT~\cite{Zhao2022ICWSM_PrivacyAlert,Lu2018NeurIPS_VilBERT} 
        & 65.80 & 69.70 & 67.70 & 89.70 & 87.90 & 88.80 & 77.75 & 83.37 & 78.80 \\
        & *GMMF~\cite{Zhao2022ICWSM_PrivacyAlert,Zhao2023TWEB} 
        & 77.90 & 72.20 & 75.00 & 91.00 & 93.20 & 92.10 & 84.45 & 87.94 & 82.70 \\
        & $\diamond$Privacy VLM~\cite{samson2024little} 
        & N/A & N/A & N/A & N/A & N/A & N/A & 54.00 & N/A & 78.00 \\
        & S2P~\cite{Xompero2025PoPETs} 
        & 63.11 & 63.11 & 63.11 & 87.67 & 87.67 & 87.67 & 75.39 & 81.51 & 75.39 \\
        & MiniCPM~\cite{Yao2024_ArXiv_MiniCPM_V} 
        & 41.34 & 67.33 & 51.23 & 86.17 & 68.05 & 76.05 & 63.75 & 67.87 & 67.69 \\
        & Phi-3-V~\cite{Abdin2024_ArXiv_Phi3} 
        & 60.69 & 42.89 & 50.26 & 82.61 & 90.71 & 86.47 & 71.65 & 78.73 & 66.80 \\ 
        & LLaVA~\cite{Liu2023_NeurIPS_LLaVA,Liu2024_CVPR_LLaVa} 
        & 41.23 & 89.33 & 56.42 & 94.15 & 57.43 & 71.34 & 67.69 & 65.42 & 73.38 \\        

        \bottomrule \addlinespace[\belowrulesep]
        \multicolumn{11}{l}{\parbox{0.97\linewidth}{\scriptsize{
        As references, we include results for the degenerate cases of predicting all images either as public or private, and for a baseline using a pseudo-random generator (Random) to sample the predictions from a uniform distribution.
        *Results taken from Zhao et al.'s work on PrivacyAlert~\cite{Zhao2022ICWSM_PrivacyAlert}. As some of the images are no longer available, performance may be higher for these methods. Unlike Zhao et al.'s work that computes the weighted average precision and recall (BA) across the two classes, giving higher emphasis to the public class that has a higher number of samples, we computed the macro averaging (unweighted mean), treating the two classes equally.
        $\diamond$Results taken from Samson et al.'s paper~\cite{samson2024little}, which reports performance measures in a different and inconsistent way compared to previous benchmarks on the dataset.
        }}}
    \end{tabular}
    \label{tab:multimodalcomparison}
\end{table*}
S2P outperforms VLMs on both PrivacyAlert and IPD in terms of overall precision, balanced accuracy, and accuracy. The multi-modal fusion combined with task-specific fine-tuning of GMMF achieves the highest accuracy and balanced accuracy on PrivacyAlert. LLaVA predicts most of the images as private, resulting in high recall at 89.33\% and 70.49\% and precision at 41.23\% and 55.79\% on PrivacyAlert and IPD, respectively. 
Despite the fine-tuning on a privacy-focused dataset, results reported for Privacy VLM show that the model does not outperform other models and potentially predicts many images as public. P-VilBERT, another fine-tuned model, predicts more images as private, achieving a higher balanced accuracy (78.80\%) than most of the other models but still lower than GMMF (82.70\%).
VLMs rely on billions of parameters and require more powerful GPUs than other privacy classifiers. With $1.30 \cdot 10^{10}$ parameters, LLaVA requires the most powerful GPU (H100) and runs at about 0.5 s/image on average. Phi-3-V and MiniCPM have fewer parameters than LLaVA, $4.20 \cdot 10^9$ and $8.50 \cdot 10^9$ respectively, requiring a less powerful GPU (RTX3090) to fit the model and thus running at 1.98$\pm$0.60 s/image and 4.60$\pm$0.99 s/image, respectively. S2P has about 24 million parameters, running at 8.75 ms/image on average on a power-efficient GPU (GTX1080).

Overall, this comparative analysis shows that large pre-trained VLMs do not yet outperform previous models for image privacy classification, and achieve lower classification performance while requiring higher computational costs. The use of different GPUs makes the speed comparison unfair. However, S2P is fast on low-memory GPU and its performance remains valid regardless of using a higher-end GPU, while the VLMs cannot be deployed on lower-memory GPUs due to their memory requirements. Although MiniCPM and Phi-3-V might be faster on more powerful GPUs, these models generate richer textual answers requiring additional post-processing and the output length could still influence their inference speed. The higher accuracy of S2P while running at a faster speed on a less powerful GPU (thanks also to the limited number of parameters) stands out and sets a relevant baseline for future comparisons. 

\newcommand{\plotTitleStrut}{\vphantom{Agjp}}
\pgfplotstableread{resources/acc_size_ipd.txt}\sizevsacc
\pgfplotstableread{resources/acc_size_privacyalert.txt}\sizevsaccprivacyalert
\begin{figure}[t!]
    \centering
    \begin{tikzpicture}
        \begin{semilogxaxis}[
            axis lines*=left,
            width=0.3\columnwidth,
            height=0.3\columnwidth,
            xlabel={\# parameters},
            ylabel={BA (\%)},
            ymin=50, ymax=85,
            xmin=7500000,
            xtick={1000000,10000000,100000000,1000000000,10000000000},
            xlabel near ticks,
            ylabel near ticks,
            label style={font=\footnotesize},
            tick label style={font=\scriptsize},
            title={\plotTitleStrut IPD},
            title style={
                font=\footnotesize,
                at={(axis description cs:0.5,0.96)},
                anchor=south
            },
            scatter/classes={
            a={mark=*,dc4},
            b={mark=*,dc5},
            c={mark=*,dc1},
            d={mark=*,dc2}
            }
        ]
        \addplot+[dashed, color=gray,mark=none] coordinates{(10000000,0) (10000000,100)};
        \addplot+[dashed, color=gray,mark=none] coordinates{(100000000,0) (100000000,100)};
        \addplot+[dashed, color=gray,mark=none] coordinates{(1000000000,0) (1000000000,100)};
        \addplot+[dashed, color=gray,mark=none] coordinates{(10000000000,0) (10000000000,100)};
        \addplot+[scatter, only marks, scatter src=explicit symbolic] table[x=size,y=ba, meta=label]{\sizevsacc};
        \end{semilogxaxis}
    \end{tikzpicture}
    \begin{tikzpicture}
        \begin{semilogxaxis}[
            axis x line*=bottom,
            hide y axis,
            width=0.3\columnwidth,
            height=0.3\columnwidth,
            xlabel={\# parameters},
            ymin=50, ymax=85,
            xmin=7500000,
            xtick={1000000,10000000,100000000,1000000000,10000000000},
            yticklabels={},
            xlabel near ticks,
            ylabel near ticks,
            label style={font=\footnotesize},
            tick label style={font=\scriptsize},
            title={\plotTitleStrut PrivacyAlert},
            title style={
                font=\footnotesize,
                at={(axis description cs:0.5,0.96)},
                anchor=south
            },
            scatter/classes={
            a={mark=*,dc4},
            b={mark=*,dc5},
            c={mark=*,dc1},
            d={mark=*,dc2}
            }
        ]
        \addplot+[dashed, color=gray,mark=none] coordinates{(10000000,0) (10000000,100)};
        \addplot+[dashed, color=gray,mark=none] coordinates{(100000000,0) (100000000,100)};
        \addplot+[dashed, color=gray,mark=none] coordinates{(1000000000,0) (1000000000,100)};
        \addplot+[dashed, color=gray,mark=none] coordinates{(10000000000,0) (10000000000,100)};
        \addplot+[scatter, only marks, scatter src=explicit symbolic] table[x=size,y=ba, meta=label]{\sizevsaccprivacyalert};
        \end{semilogxaxis}
    \end{tikzpicture}
    \hspace{0.115cm}
    \begin{tikzpicture}
        \begin{semilogxaxis}[
            hide y axis,
            axis x line*=bottom,
            width=0.3\columnwidth,
            height=0.3\columnwidth,
            xlabel={s/image},
            ymin=50, ymax=85,
            xtick={0.001,0.01,0.1,1,10},
            xticklabels={,,0.1,1,10},
            yticklabels={},
            xlabel near ticks,
            ylabel near ticks,
            label style={font=\footnotesize},
            tick label style={font=\scriptsize},
            title={\plotTitleStrut IPD},
            title style={
                font=\footnotesize,
                at={(axis description cs:0.5,0.98)},
                anchor=south
            },
            scatter/classes={
            a={mark=*,dc4},
            b={mark=*,dc5},
            c={mark=*,dc1},
            d={mark=*,dc2}
            }
        ]
        \addplot+[dashed, color=gray,mark=none] coordinates{(0.01,0) (0.01,100)};
        \addplot+[dashed, color=gray,mark=none] coordinates{(0.1,0) (0.1,100)};
        \addplot+[dashed, color=gray,mark=none] coordinates{(1,0) (1,100)};
        \addplot+[scatter, only marks, scatter src=explicit symbolic] table[x=speed-m,y=ba, meta=label]{\sizevsacc};
        \end{semilogxaxis}
    \end{tikzpicture}
    \hspace{0.275cm}
    \begin{tikzpicture}
        \begin{semilogxaxis}[
            axis x line*=bottom,
            hide y axis,
            width=0.3\columnwidth,
            height=0.3\columnwidth,
            xlabel={s/image},
            ymin=50, ymax=85,
            xtick={0.001,0.01,0.1,1,10},
            xticklabels={,,0.1,1,10},
            yticklabels={},
            xlabel near ticks,
            ylabel near ticks,
            label style={font=\footnotesize},
            tick label style={font=\scriptsize},
            title={\plotTitleStrut PrivacyAlert},
            title style={
                font=\footnotesize,
                at={(axis description cs:0.5,0.98)},
                anchor=south
            },
            scatter/classes={
                a={mark=*,dc4},
                b={mark=*,dc5},
                c={mark=*,dc1},
                d={mark=*,dc2}
            }
        ]
            \addplot+[dashed, color=gray,mark=none] coordinates{(0.01,0) (0.01,100)};
            \addplot+[dashed, color=gray,mark=none] coordinates{(0.1,0) (0.1,100)};
            \addplot+[dashed, color=gray,mark=none] coordinates{(1,0) (1,100)};
            \addplot+[scatter, only marks, scatter src=explicit symbolic] table[x=speed-m,y=ba, meta=label] {\sizevsaccprivacyalert};
        \end{semilogxaxis}
    \end{tikzpicture}
    \caption{Comparison of 
    \protect\tikz \protect\fill[dc1,fill=dc4] (1,1) circle (0.5ex);~Phi-3-V,
    \protect\tikz \protect\fill[dc2,fill=dc5] (1,1) circle (0.5ex);~MiniCPM,
    \protect\tikz \protect\fill[dc5,fill=dc1] (1,1) circle (0.5ex);~LLaVA,
    \protect\tikz \protect\fill[dc4,fill=dc2] (1,1) circle (0.5ex);~S2P
    in terms of balanced accuracy (BA) versus number of parameters (\# parameters) and inference speed (s/image). Best performance on the top-left corner. 
    The x-axis uses a logarithmic scale.    
    }
    \label{fig:sizevsacc}
\end{figure}

\section{Robustness to Image Perturbations} 

We analyse the robustness of LLaVA, Phi-3-V, and S2P to multiple image perturbations, such as compression, changes in illumination, and the addition of noise, on the testing set of PrivacyAlert~\cite{Zhao2022ICWSM_PrivacyAlert}. MiniCPM is excluded due to the lower performance compared to LLaVA and Phi-3-V. Moreover, we consider the best-performing case: LLaVA with P1 and Phi-3-V with P1 and the conservative answer-to-decision mapping approach. 

\input{tex/robustness_comp_icpr}

We use lossy JPEG compression by varying the quality parameter in the interval $[0,100]$, where the higher the value, the better the visual quality. We choose the following quality values: $\{100, 99, 95, 85, 75, 50, 25\}$. We use a step of 5 between 100 and 75, and also include 99, as quality values to analyse the robustness to small compression effects. Note that the value 100 should preserve the original image quality, however the encoding process can still influence the image and hence the model performance.
For \textit{illumination changes}, we modify the brightness of an image by varying the gamma correction parameter. As gamma correction is a non-linear transformation depending on each pixel intensity, we use the following values (inverted with respect to the central value 1, i.e. no change in illumination): $\{0.1, 0.4, 0.67, 1, 1.5, 2.5, 10\}$. As \textit{noise}, we add pseudo-random zero-mean Gaussian noise with standard deviation varying with the following values: $\{0, 0.01, 0.05, 0.10, 0.20, 0.30, 0.40\}$; and pseudo-random salt noise varying the threshold with the following values: $\{255, 245, 225, 200, 175, 150\}$. 
Applying the 0 value for zero-mean Gaussian noise and the value 255 for salt noise corresponds to no perturbation applied to the original image. We apply all these perturbations before re-scaling and normalising the images to use as input to the models. 

Figure~\ref{fig:imageperturcompression} shows that LLaVA is the most robust to the perturbations, especially for JPEG compression, Gaussian noise, and illumination changes.  
On the contrary, the model's balanced accuracy decreases under salt noise due to more images predicted as private, thus resulting in a higher recall for the private but lower recall for the public class. 
Phi-3-V is also robust to JPEG compression and illumination changes but more sensitive to noise than LLaVA. The added perturbations, except for JPEG compression, increase the percentage of images that require mapping the ambiguous answer to the final decision from 7.07\% (127 images), when no perturbation is added to the image, to the worst case of 33.46\% (601 images) when applying Gaussian noise with a standard deviation of 0.4. The conservative approach (answers set to private class) influences the analysis of the classification performance, resulting in higher robustness, but the quantification of the mapped answers shows the higher sensitivity of Phi-3-V to illumination changes and noise.

\begin{figure}[t!]
    \centering 
    \scriptsize
    \setlength\tabcolsep{2pt}
    \includegraphics[width=0.99\columnwidth]{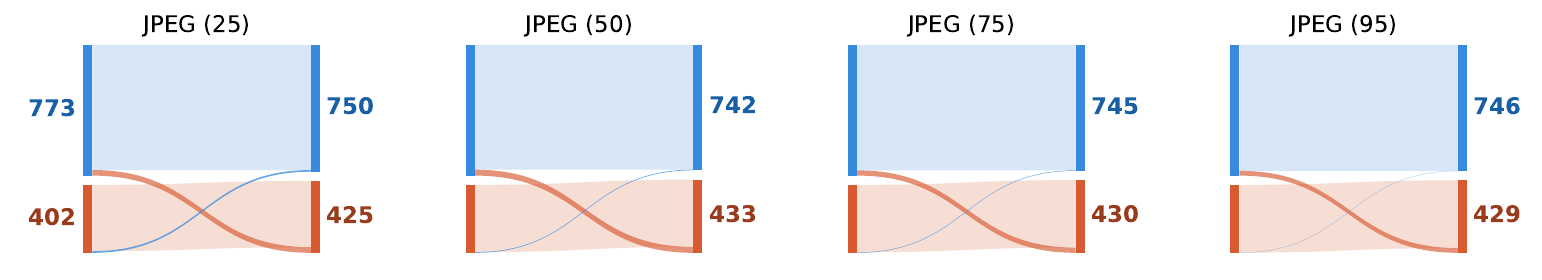}\\
    \vspace{-5pt}
    \includegraphics[width=0.99\columnwidth]{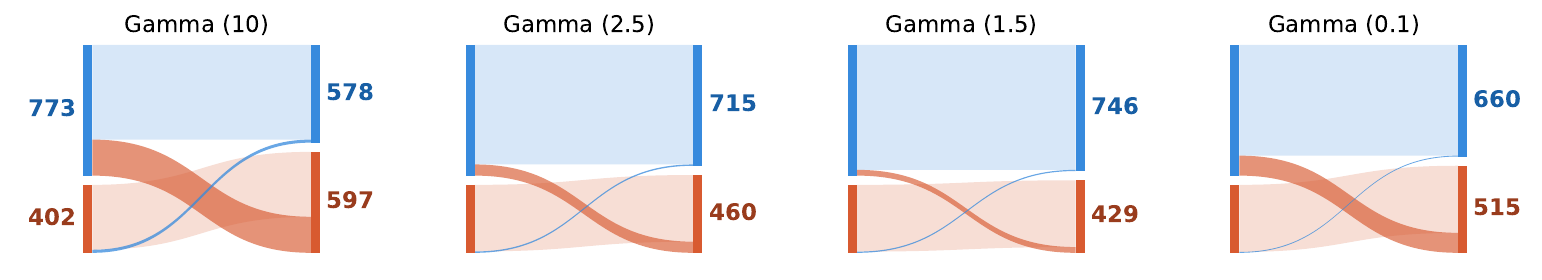}\\
    \vspace{-5pt}
    \includegraphics[width=0.99\columnwidth]{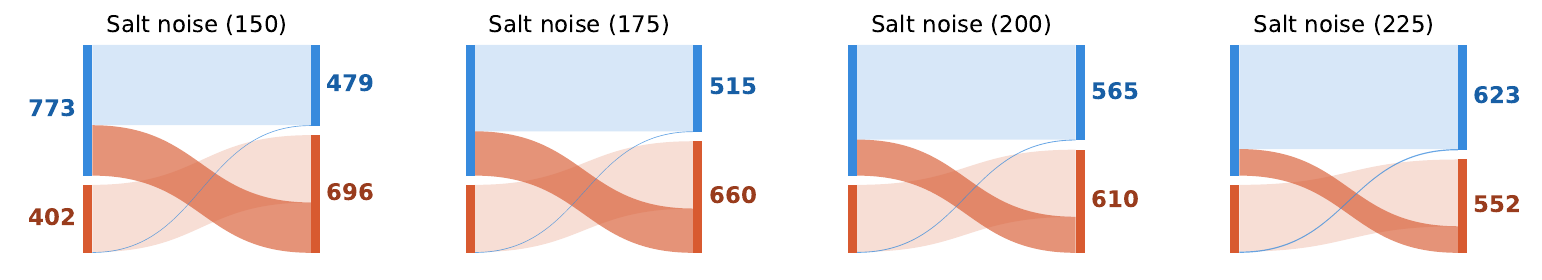}\\
    \vspace{-5pt}
    \includegraphics[width=0.99\columnwidth]{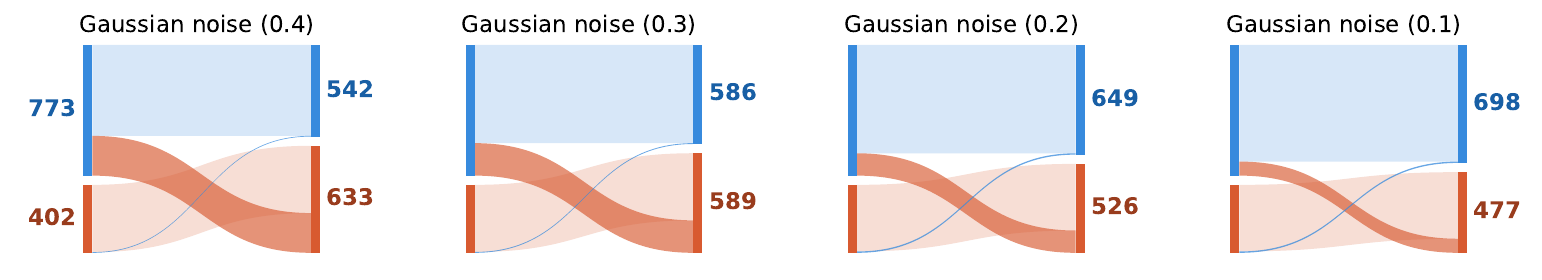}\\

    \caption{Changes in privacy labels predicted by LLaVA when perturbing images from the PrivacyAlert testing set, with different degradation intensities for JPEG compression (quality), illumination changes (gamma value), salt noise, and Gaussian noise (standard deviation). \textit{Left side of each plot}: number of images correctly classified as public (\protect\tikz \protect\fill[dc23,fill=dc23] (1,1) circle (0.5ex);) or private (\protect\tikz \protect\fill[dc24,fill=dc24] (1,1) circle (0.5ex);) before applying any perturbations.  \textit{Right side of each plot}: number of images predicted as private or public after applying any perturbations. The thinner the lines connecting flipped labels, the more robust the model to the perturbation.  }
    \label{fig:sankey}
  \end{figure}

Figure~\ref{fig:sankey} analyses the prediction flips of correctly classified clean images by LLaVA across the four perturbation types. Aggregate metrics such as balanced accuracy do not reveal whether errors create privacy risks or affect model usability, since flipping private images to public can expose sensitive content, while flipping public images to private makes the system more conservative. Analysing prediction flips helps understand the privacy risks from safer over-protection. Limitations also apply to the recall for the private class, where two opposite flips can preserve the score while hiding the real robustness. Changes in predicted labels are mainly one-directional with public images classified as private, while private predictions are highly stable. This shows that images correctly classified as private on the clean input remain mostly private under the tested perturbations. From a privacy-protection perspective, this is a positive behaviour since private content remains flagged as private even when the input is perturbed. 

\begin{figure}[t!]
\centering
\scriptsize
\setlength\tabcolsep{11pt}
\renewcommand{\arraystretch}{1.2}
\begin{tabular}{m{0.13\linewidth} m{0.75\linewidth}}
\includegraphics[height=1.1\linewidth]{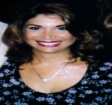} & 
\color{black} Original image. \color{black} \newline
``The image shows a person wearing a floral patterned top and a necklace, $<$the content of the image does not provide any information regarding private or sensitive data$>$''.  \\[6pt]

\includegraphics[height=1.1\linewidth]{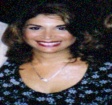} &
\color{black} Salt noise value: 245. \color{black} \newline
``Yes, the image is likely to contain private information as it shows a person's face, which is typically considered sensitive information.''  \\[6pt]

\includegraphics[height=1.1\linewidth]{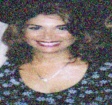} &
\color{black} Salt noise value: 200. \color{black}  \newline
``Unable to determine from the image provided.'' \\[6pt]

\includegraphics[height=1.1\linewidth]{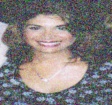} &
\color{black} Salt noise value: 175. \color{black} \newline
``The image shows a person wearing a patterned garment with a visible neckline, but due to the pixelation, it is not possible to provide a detailed description or identify the individual. The question asks for an assessment of the image's content regarding private information, which cannot be accurately determined from the provided section of the image.'' 
\\[6pt]

\includegraphics[height=1.1\linewidth]{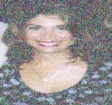} & 
\color{black}  Salt noise value: 150. \color{black} \newline
``The image appears to be a low-resolution, pixelated version of a person, which does not provide any specific information about the individual's identity or privacy. However, without knowing the context or the source of the image, it is not possible to determine if it contains private information.'' 
\\[6pt]

\end{tabular}
\caption{Phi-3-V answers when varying the pseudo-random salt noise from top to bottom added to a sample private image from PrivacyAlert~\cite{Zhao2022ICWSM_PrivacyAlert}. Note that images are resized.}
\label{fig:vizresrobustness}
\end{figure}

Figure~\ref{fig:vizresrobustness} shows an example of a private image perturbed with various levels of salt noise and the corresponding answers from Phi-3-V. Answers substantially vary when increasing the perturbation, switching from a public preference (but set to private due to the lack of ``[Yes]'' or ``[No]'' in the answer) to an undetermined answer, and becoming more descriptive.
Finally, S2P is robust to JPEG compression but the performance decreases under heavy salt and Gaussian noises, and large illumination changes (i.e. darker image). Training with data augmentation could make S2P more robust to these perturbations.

\section{Conclusion}
\label{sec:conclusion}

We defined a protocol for the systematic evaluation of the top-3 open-source instruction-following VLMs for zero-shot private image classification on standard datasets, such as PrivacyAlert~\cite{Zhao2022ICWSM_PrivacyAlert} and IPD~\cite{Yang2020PR}. Our evaluation includes the robustness to common image perturbations that can affect the decision of a model and incorporates prior work to compare the general-purpose models with previous vision-only and multi-modal models designed and trained for image privacy. Although some VLMs (e.g. LLaVA) showed stronger robustness to perturbations, they achieved lower balanced accuracy than task-specific models and require significantly more computational resources.

Future work will include the design of small vision-language model architectures and fine-tuning strategies tailored to privacy, broader prompt engineering and VLMs evaluation, the assessment against adversarial noise, and integration of visual privacy classification within text-based document processing systems.

\small
\subsubsection{Acknowledgements.}   This work was partially supported by the CHIST-ERA programme through the project GraphNEx (CHIST-ERA-19-XAI-006), under UK EPSRC grant EP/V062107/1, Swiss  NSF grant 195579, and France ANR grant ANR-21-CHR4-0009. 
A. Xompero did part of the work when he was affiliated with Queen Mary University of London.
A.~E.~Baia did part of the work when she was affiliated with the Idiap Research Institute.

\subsubsection{Disclosure of Interests.} The authors have no competing interests to declare that are relevant to the content of this article.

%
%
%
\bibliographystyle{splncs04}
\bibliography{main}

\end{document}